\pdfoutput=1
\documentclass{article}

    \usepackage[preprint, nonatbib]{nips_2018}
    
    \usepackage[utf8]{inputenc} 
    \usepackage[T1]{fontenc}    
    \usepackage{hyperref}       
    \usepackage{url}            
    \usepackage{booktabs}       
    \usepackage{amsfonts}       
    \usepackage{nicefrac}       
    \usepackage{microtype}      
    \usepackage{amsfonts}
    \usepackage{amsmath}

    \usepackage{graphicx}
    \usepackage{xcolor}
    \usepackage{import}
    \usepackage{transparent}

    \title{Seq2Seq Mimic Games:\\ A Signaling Perspective}
    
    \author{
	  Juan Leni\\
      University of Strathclyde\\
      \texttt{juan.leni@strath.ac.uk} \\
      \And
      John Levine\\
	  University of Strathclyde \\
      \texttt{john.levine@strath.ac.uk} \\
      \AND
      John Quigley \\
	  University of Strathclyde \\
	  \texttt{j.quigley@strath.ac.uk}
    }
  
\begin{document}
\graphicspath{{figs/}}

\maketitle

\begin{abstract}
	We study the emergence of communication in multiagent adversarial settings inspired by the classic Imitation game.
	A class of three player games is used to explore how agents based on sequence to sequence (Seq2Seq) models can learn to communicate information in adversarial settings.
    We propose a modeling approach, an initial set of experiments and use signaling theory to support our analysis.
    In addition, we describe how we operationalize the learning process of actor-critic Seq2Seq based agents in these communicational games. 
\end{abstract}

\section{Introduction}
We propose an initial step towards models to study the emergence of communication in adversarial environments. In particular, we explore Seq2Seq \cite{sutskever_sequence_2014}\cite{cho_learning_2014}\cite{vinyals_neural_2015} based agents in a class of games inspired by the Imitation game \cite{turing_computing_1950}. 

We analyze these games from the perspective of signaling games \cite{spence_job_1973}\cite{zahavi_mate_1975}. Agents are required to learn how to maximize their expected reward by improving their communication policies. Experiments do not assume previous knowledge or training and all agents are a priori ungrounded, i.e. tabula rasa.

Training Seq2Seq models is known to be challenging \cite{bachman_data_2015}\cite{bahdanau_actor-critic_2016}\cite{yu_seqgan_2016}\cite{havrylov_emergence_2017}. In this work, we use an actor-critic architecture, however unlike Bahdanau et al.\cite{bahdanau_actor-critic_2016}, our emphasis is not on sequence prediction but on maximizing expected rewards by developing a adequate communicational strategy.

When analyzed from a signaling perspective, we show how agents in adversarial conditions may learn to communicate and transfer information when intrinsic or environmental conditions are adequate: e.g. handicap and computational advantages. Depending on the game structure, we show how agents reach separating or pooling equilibria \cite{spence_job_1973}. 

\subsection{Game class description}
We study a class of three-player imperfect information iterated games. Two agents, one of each
type $c \in \{blue, red\}$ (i.e. $\mathcal{A}_{blue}$ and $\mathcal{A}_{red}$), and a single
interrogator agent ($\mathcal{I}$) exchange messages $m$, i.e. sequences $m=s_1s_2{\cdots}s_L$
of discrete symbols $s_l \in \Sigma$ from a prearranged alphabet $\Sigma$. This alphabet includes a special symbol <EOS> to indicate the end of sentences. The lack of common knowledge is a key distinction with respect to the classic Imitation game definition. Agents are not aware of their own or others limitations. Like classic reinforcement learning, they need to explore and discover their competitive advantages or disadvantages through interaction.

In every iteration $t$, the interrogator $\mathcal{I}$ starts by sending a \textit{question/primer} $m_t^Q$ to both $\mathcal{A}_c$ agents. Afterwards, $\mathcal{I}$ receives their answers as two anonymous messages $m_t^{A_i}$ and is rewarded if it can determine the type $c_i$ of the source of each message. $Blue$ and $red$ are in adversarial positions. While $\mathcal{A}_{blue}$ is rewarded when its messages are recognized, $\mathcal{A}_{red}$ is rewarded when it misleads $\mathcal{I}$ and ${c=red}$ messages are incorrectly identified as $c=blue$. After each iteration, all messages, sources, rewards and $\mathcal{I}$'s inferred types are made available to all agents. Agents use this information to update their communication strategies and beliefs.


\section{Related Work}
Attention has been recently brought to the communicational aspects of multiagent systems. Recent research has explored cooperative \cite{foerster_learning_2016}\cite{lazaridou_multi-agent_2016}\cite{havrylov_emergence_2017} and semi-cooperative scenarios such as negotiation games \cite{cao_emergent_2018}. The emergence of communication in adversarial scenarios has been explored less extensively.

A variation of the classic Imitation game \cite{turing_computing_1950}, the famous Turing test, has been considered for many years a canonical proof of intelligent behavior. While the validity of this assertion is controversial \cite{pinarsaygin_turing_2000}, the game definition is very valuable. There is a significant overlap with signaling theory \cite{spence_job_1973} and signaling games have been extensively explored in the economics and game theory literature.

While our motivation is to study the emergence of communication as a signaling mechanism, we notice how Kannan et al.\cite{kannan_adversarial_2017} use a related approach to evaluate model performance. In their work, they propose a generative adversarial network (GAN) \cite{goodfellow_generative_2014} where a discriminator had to differentiate between real sentences and a generative model. 

Generative adversarial networks (GANs) \cite{goodfellow_generative_2014} have resulted in a wide range of interesting adversarial applications. However, the extension to sequence to sequence models (Seq2Seq) \cite{sutskever_sequence_2014} has been difficult. Combining GAN with Seq2Seq models is challenging because discrete samples drawn from categorical distributions hinder backpropagation. In addition, alternatives on how to perform reward imputation to partial sequences \cite{bachman_data_2015} have been proposed.

With respect to backpropagating errors, reparametrization \cite{graves_stochastic_2016} has been used multiple times to allow for backpropagation through stochastic nodes. In particular, Gumbel-Softmax\cite{kusner_gans_2016} has allowed categorical distributions \cite{jang_categorical_2016} in stochastic computational graphs. This technique has been shown as an alternative to reinforcement learning \cite{havrylov_emergence_2017} within the scope of cooperative referential games. More recently, similar ideas resulted in SeqGAN \cite{yu_seqgan_2016} being proposed. Further incremental improvements have been published, such as applying actor-critic models \cite{bahdanau_actor-critic_2016} or combining with proximal policy optimization (PPO) \cite{tuan_proximal_2018} in order to improve learning performance.


\section{Model}
We use Seq2Seq actor-critic models in every agent. While actors are parametric generative models that produce sequences, critics provide a subjective estimation of the expected reward for a given partial sequence. Agents train a critic using data and later optimize their behavioral strategy (actor) using the critic's feedback. After each round, all information is made public, so critics can be trained using all available experiences including adversaries' responses.

\begin{figure}[h]
	\centering
	\def\svgwidth{13cm}
	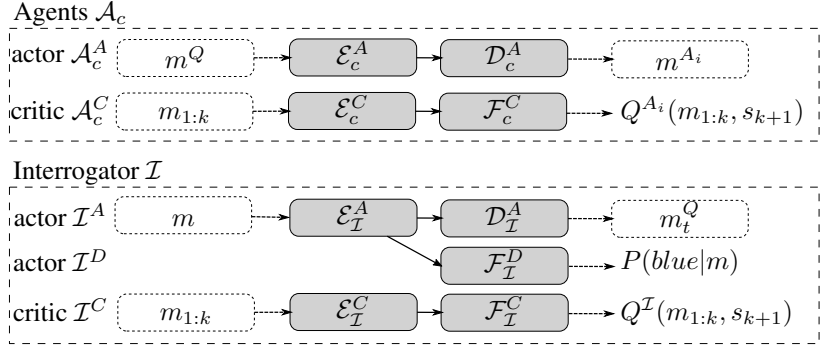
	\caption{Agent models}\label{fig:models}
\end{figure}

Figure \ref{fig:models} shows a block diagram describing the agents' model structure. All encoders and decoders use gated recurrent units (GRU) \cite{cho_learning_2014} and a simple attention mechanism \cite{bahdanau_neural_2014}. To simplify notation, we omit subscript $t$ unless the context is not clear.

\paragraph{Actors} $\mathcal{A}_c$ agents ($c \in \{red, blue\}$) use a Seq2Seq model \cite{sutskever_sequence_2014} as actor, i.e. an GRU encoder $\mathcal{E}_c^A$ followed by a GRU decoder $\mathcal{D}_c^A$ that terminates in a softmax. 
The interrogator $\mathcal{I}$ actor has two output branches: an encoder $\mathcal{E}_\mathcal{I}^A$ followed by a decoder $\mathcal{D}_\mathcal{I}^A$ or a discriminator $\mathcal{C}_\mathcal{I}^A$. Which output is used depends on the game step (questioning vs classifying).

\paragraph{Critics} Critics follow a similar structure: a GRU encoder $\mathcal{E}^C$ followed by a feed-forward network $\mathcal{F}^C$ that terminates in a softmax function. Given a partial input sequence $m_{1:k}=s_1 {\cdots} s_k$, the critic estimates the corresponding q-value $Q(m_{1:k}, s_{k+1})$ for every possible $s_{k+1} \in \Sigma$. We use a technique similar to DQN \cite{mnih_human-level_2015} and vectorize the calculation to obtain a single vector with $Q$ values for each possible $s_{k+1} \in \Sigma$.

\subsection{Learning}
In every iteration $t$, we sample $N$ end-to-end interactions $(m^Q, c^{A_1}, c^{A_2}, m^{A_1}, m^{A_2}, c_1^\mathcal{I}, c_2^\mathcal{I} )$ where $c_i^{\mathcal{I}}$ indicates the inferred type with respect to message $m^{A_i}$. There are two training stages. We first train critics and the discriminator using sampled data. Secondly, we train actors using feedback from their respective critics.

\paragraph{Training with data} Samples are used to train both the critics and the discriminator. All critics ($\mathcal{A}_c^C$, $\mathcal{I}^C$) are trained using concatenated sequences $m=m^Q||m^{A_i}$ as input. We use binary cross entropy $H(x, y) = - \left[ y \cdot \log x + (1 - y) \cdot \log (1 - x) \right]$ and target values $\eta(x) = \delta_{c^{\mathcal{I}}, x}$ where $\delta_{i,j}$ is the Kronecker delta. 

The corresponding losses for each message $m$ and respective known type $c$ are shown in Table \ref{tbl:losses1}. While critics are trained using partial sequences, the discriminator only considers full messages. Losses are accumulated over all samples and optimization is done using Adam \cite{kingma_adam_2014}.

\begin{table}[htb]
	\caption{Critics and Discriminator Losses for each sample} \label{tbl:losses1}
	\centering
	\begin{tabular}{ccc}
		\toprule
		\cmidrule(r){1-2}
		$\mathcal{L}_{\mathcal{A}_c}^{C}(m, c)$  & $\mathcal{L}_{\mathcal{I}}^{C}(m, c)$ & $\mathcal{L}_{\mathcal{I}}^{D}(m, c)$ \\
		\midrule
		
		\(\displaystyle \sum_{k=1}^{|m|}  H\Big[Q^{\mathcal{A}_c}(m_{:k-1}, m_k) , \eta(blue) \Big]    \) &
		\(\displaystyle \sum_{k=1}^{|m|}  H\Big[Q^{\mathcal{I}}(m_{:k-1}, m_k)   , \eta(c)    \Big]    \) &
		\(\displaystyle                   H\Big[\mathcal{I}^D(m), \eta(c)                     \Big]    \) \\
		\bottomrule
	\end{tabular}
\end{table}

\paragraph{Training with critics} We train actors using feedback from their respective critics. We use trigger messages to obtain a response from actors, i.e. $\mathcal{A}^A_c$, receives $m^Q$ and outputs $m^{A_i}$. We use $m^Q$ messages from previous training samples. For the interrogator, we use fixed empty triggers, actor $\mathcal{I}^A$, receives $m^T=\text{EOS}$ and outputs $m^Q$. As each symbol $m_k$ is generated, we retrieve both the symbol and the corresponding multinomial distribution $\pi_{m_{1:k-1}}$ used by the decoder. 

From the respective critic, we obtain a vector $Q(m_{1:k-1}, *)$ with q-values for each possible symbol. The dot product $\pi_{m_{1:k-1}} \cdot Q(m_{1:k-1}, *)$ results in the policy expected reward. The sum of the scores of all samples is later optimized using Adam \cite{kingma_adam_2014}.

\section{Experiments}
We ran an initial set of experiments where we varied environmental properties or agent capabilities. All experiments involve continuous learning. Experiment parameters are shown in Table \ref{tbl:experiments}. 

Within these experiments, it is important to remember that it is in the best interest of agent $red$ not to reveal its identity. In the games we explore, two kinds of equilibria are expected: separating or pooling. In pooling equilibria, the interrogator is not able to extract enough information from messages to determine sources correctly. Instead, separating equilibria is reached whenever messages carry enough information for interrogator to effectively determine the source.

\begin{table}[htb]
	\caption{Experiments} \label{tbl:experiments}
	\centering
	\begin{tabular}{lcclllc}
		\toprule
		\cmidrule(r){1-2}
		Experiment          & $\mathcal{I}$ & $\mathcal{A}_{blue}$  & $\mathcal{A}_{red}$   & Outcome        \\
		\midrule
		1 Identical      & $h=8$ & $h=8$, $|m^{A}|\leq8$ & $h=8$, $|m^{A}|\leq8$ & pooling        \\
		2 Handicap blue  & $h=8$ & $h=8$, $|m^{A}|\leq6$ & $h=8$, $|m^{A}|\leq5$ & separating     \\
		3 Handicap red   & $h=8$ & $h=8$, $|m^{A}|\leq5$ & $h=8$, $|m^{A}|\leq6$ & pooling        \\
		\midrule
		4 Neurons A blue & $h=8$ & $h=8$, $|m^{A}|\leq8$ & $h=4$, $|m^{A}|\leq8$ & separating     \\
		5 Neurons A red  & $h=8$ & $h=4$, $|m^{A}|\leq8$ & $h=8$, $|m^{A}|\leq8$ & pooling        \\
		6 Neurons B blue & $h=8$ & $h=16$, $|m^{A}|\leq8$ & $h=8$, $|m^{A}|\leq8$ & pooling      \\
		7 Neurons B red  & $h=8$ & $h=8$, $|m^{A}|\leq8$ & $h=16$, $|m^{A}|\leq8$ & pooling      \\
		\bottomrule
		\multicolumn{5}{l}{$h$ indicates the number of hidden units in the encoder/decoder}
	\end{tabular}
\end{table}

We divide experiments in two groups. In the first group (experiments 1 to 3), we disable questioning by limiting the number of symbols to zero. The second group (experiments 4 to 7) allows for questioning/priming. 

In all games, we impose limitations on the number of Seq2Seq hidden units $h$ and the maximum amount of symbols an agent can emit before their communication channel is terminated. Agents are not explicitly aware of these factors but can indirectly perceive them through interaction.

The \textit{Identical} experiment is a game where all settings and $\mathcal{S}_c$ agents are identical. The interrogator has no chance of differentiating them resulting as expected in a pooling equilibrium. We use this game to confirm basic behavior.

In \textit{Handicap} experiments (2 and 3), we assign different length limits to agents. Experiment 2 shows how $blue$ discovers this advantage to distinguish itself (separating equilibrium). However, when the advantage is given to $red$ (\textit{experiment 3}), it is in $red$ best interest to hide this fact and the game converges to a pooling equilibrium.

The second group of experiments enables questioning. We vary the amount of hidden units in the encoder/decoders as described in Table \ref{tbl:experiments}. 

Results of experiment 5 are expected: $red$ does not exploit its advantages and is able to mimic $blue$ leading to a pooling equilibria. 
Experiment 4 shows an interesting outcome. The interrogator is able to separate types, we intuitively expected $\mathcal{I}$ to randomize questions, instead both $\mathcal{I}$ and $blue$ converge to fixed sequences. Detailed analysis shows that due the limited amount of hidden units, $red$ is not able to mimic the constant outputs of $blue$. Stochastic gradient descent updates affect already learned responses. This results in unstable outputs that allow $\mathcal{I}$ to differentiate between agents. When the number of hidden units increases (\textit{Neurons B} (6, 7)), the issue does not occur anymore. For a higher number of units, we could not detect situations where the interrogator was able to different sources.

\section{Conclusions}
In this work, we presented a broad class of games inspired on the Imitation game that we relate to signaling theory. We used these games to explore how communication may arise in adversarial scenarios. We explored some factors that may enable or hinder separating equilibria. To our knowledge, this is the first piece of research that explores signaling theory in games that involve Seq2Seq models. Last but not least, we present a simple operational approach to train ungrounded Seq2Seq agents in this domain. In future work, we intend to pre-training agents in some known language such as English. This will allow us to explore a more complex range of experiments by extending our work to question-answering settings and grounded communication.

\newpage
\small
\bibliography{nips_2018}
\bibliographystyle{plain}
\end{document}